%% file: main.tex
\documentclass[runningheads]{format/llncs}

\input{packages}

\begin{document}

\input{commands}

\title{AI-driven visual monitoring\\ of industrial assembly tasks}
\author{Mattia Nardon\inst{1} \and Stefano Messelodi\inst{1} \and Antonio Granata\inst{2} \and\\ Fabio Poiesi\inst{1} \and Alberto Danese\inst{2} \and Davide Boscaini\inst{1}}
\authorrunning{M. Nardon et al.}
\institute{Fondazione Bruno Kessler, Trento, Italy \and Meccanica del Sarca s.p.a., Trento, Italy}

\maketitle

\input{sections/0_abstract}

\input{sections/1_intro}
\input{sections/2_related}

\input{sections/3_method}
\input{sections/4_results}
\input{sections/5_conclusion}

{
\bibliographystyle{format/splncs04}
\bibliography{main}
}

\end{document}

%% file: packages.tex
\usepackage[T1]{fontenc}
\usepackage{graphicx}

\usepackage[english]{babel}
\usepackage{amsmath}
\usepackage{booktabs}
\usepackage{microtype}
\usepackage{xspace}
\usepackage{overpic}
\usepackage{todonotes}

\usepackage{multirow}
\usepackage{pifont}

\usepackage[colorlinks=true, allcolors=blue]{hyperref}

\usepackage{subcaption}
\usepackage{amsfonts}

%% file: commands.tex
\newcommand{\acronym}{ViMAT\xspace}
\newcommand{\lego}{LEGO\xspace}

\newcommand{\cmark}{\ding{51}}
\newcommand{\xmark}{\ding{55}}

\newcommand{\eg}{e.g.,\xspace}
\newcommand{\ie}{i.e.,\xspace}

\definecolor{forestgreen}{RGB}{0,174,88}
\definecolor{bluediagram}{RGB}{0, 102, 204}
\definecolor{graydiagram}{RGB}{102, 102, 102}
\definecolor{reddiagram}{RGB}{255, 0, 0}
\definecolor{tableazure}{RGB}{214, 234, 248}

\newcommand{\warning}[1]{\textbf{\color{red!90}{#1}}}
\newcommand{\mattia}[1]{\todo[color=blue!20, inline, author=Mattia]{#1}}
\newcommand{\stefano}[1]{\todo[color=red!20, inline, author=Stefano]{#1}}
\newcommand{\davide}[1]{\todo[color=yellow!20, inline, author=Davide]{#1}}

%% file: sections/0_abstract.tex
\begin{abstract}
Visual monitoring of industrial assembly tasks is critical for preventing equipment damage due to procedural errors and ensuring worker safety.
Although commercial solutions exist, they typically require rigid workspace setups or the application of visual markers to simplify the problem.
We introduce \acronym, a novel AI-driven system for real-time visual monitoring of assembly tasks that operates without these constraints.
\acronym combines a perception module that extracts visual observations from multi-view video streams with a reasoning module that infers the most likely action being performed based on the observed assembly state and prior task knowledge.
We validate \acronym on two assembly tasks, involving the replacement of \lego components and the reconfiguration of hydraulic press molds, demonstrating its effectiveness through quantitative and qualitative analysis in challenging real-world scenarios characterized by partial and uncertain visual observations.
Project page: \url{https://tev-fbk.github.io/ViMAT}.

\keywords{Visual monitoring \and Assembly task \and Object detection.}
\end{abstract}

%% file: sections/1_intro.tex
\section{Introduction}\label{sec:intro}

\input{figures/teaser/teaser}

Assembly tasks, involving the procedural integration of atomic components to achieve a target configuration, are central to manufacturing~\cite{urgo2019human}.
In this work, we focus on the reconfiguration of hydraulic press molds, a critical process required when switching production between different products.
Improper mold reconfiguration can damage the mold or the entire hydraulic press, incurring high repair costs and halting production for weeks or even months.
Beyond financial loss, incorrectly assembled components can be ejected at high speed during molding, posing serious (potentially fatal) safety risks to nearby workers.
As such, monitoring human-operated assembly tasks from video streams is essential.
However, this is a challenging problem due to the partial visibility of both the worker's actions and the manipulated components, often caused by inherent occlusions.
Industrial environments further complicate the task with cluttered workspaces and varying lighting conditions.
To address these challenges, existing commercial solutions~\cite{csem} often simplify the problem by imposing rigid workspace setups, where assembly components are placed in predefined containers based on their category or role, or by applying visual markers to objects, containers, and the worker's wrists to facilitate action recognition.
However, these approaches restrict worker flexibility, fail to generalize to unstructured work environments, and are prone to failure in real-world industrial settings, where visual markers can be easily compromised by dirt or debris.

In this work, we introduce \acronym, a novel AI-driven system for real-time \underline{Vi}sual \underline{M}onitoring of industrial \underline{A}ssembly \underline{T}asks that operates without such constraints.
We formalize the problem as a state-transition system, where states are specific configurations of assembly components, and transitions between them occur depending on the actions performed by the human operator.
\acronym combines a perception module, which extracts visual observations from multi-view video streams, with a probabilistic reasoning module that infers the most likely action being performed based on the observed assembly state and prior task knowledge (Fig.~\ref{fig:teaser}).
The perception module leverages a deep learning-based object detector trained exclusively on synthetic data generated from digital twins that replicate the physical assembly components and eliminate the need for manual annotation of real-world data.
The effectiveness of the resulting detector eliminates the need for rigid workspace organization and visual markers, while its robustness to lighting variations enables accurate detection in real-world industrial settings.
To mitigate occlusions, we capture video footage from multiple viewpoints, perform object detection independently for each viewpoint, and then fuse the resulting predictions to produce spatially coherent observations.
The probabilistic reasoner uses these observations to predict the full sequence of actions leading to the current configuration and demonstrates robustness to ambiguous actions.

We validate \acronym on two distinct assembly tasks: replacing components in a \lego-based system and reconfiguring a hydraulic press.
Although both involve substituting certain components with others, the former is performed in a more controlled environment with simple \lego bricks, whereas the latter occurs in a real manufacturing facility with actual mechanical components.
In both scenarios, \acronym achieves accurate real-time visual monitoring despite the challenges posed by real-world conditions.
A comprehensive evaluation, including both quantitative and qualitative results, demonstrates the effectiveness of the proposed system and its advantages over baseline approaches.

In summary, our main contributions are:
\begin{itemize}
    \item We formalize the problem of visual monitoring of assembly tasks as a state-transition system that infers the action being performed from partial visual observations on the assembly state and prior task knowledge.
    \item We enhance \acronym's perception with a deep learning-based object detector trained exclusively on synthetic data generated from task-specific digital twins, eliminating the need for real-world annotations.
    \item We validate \acronym on two distinct real-world assembly scenarios, demonstrating accurate real-time visual monitoring without relying on rigid constraints.
\end{itemize}

%% file: figures/teaser/teaser.tex
\begin{figure}[t]
    \centering
    \begin{overpic}[trim=0 0 0 0, clip, width=\textwidth]{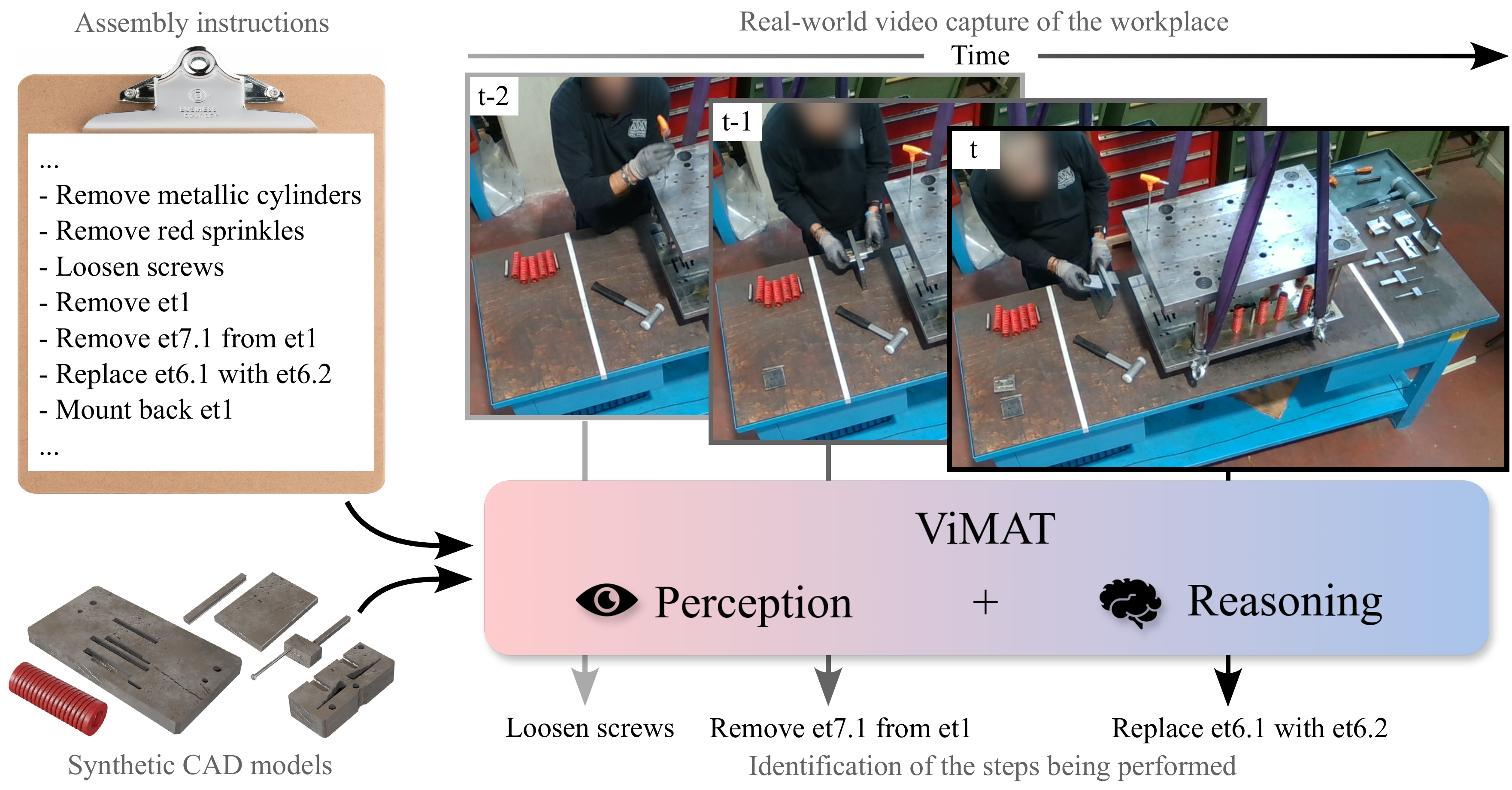}
    \end{overpic}

    \vspace{-3mm}
    \caption{
    We present \acronym, a novel system for the real-time visual monitoring of industrial assembly tasks. Given prior knowledge on assembly instructions (top left) and synthetic CAD models of assembly components (bottom left), \acronym integrates an AI-driven perception module, which extracts visual observations from real-world video streams (top right), with a probabilistic reasoning module that predicts the assembly state from these observations (bottom center-right).
    }
    \label{fig:teaser}
\end{figure}

%% file: sections/2_related.tex
\section{Related works}\label{sec:related}

\noindent\textbf{Visual monitoring} of human-operated tasks aims to identify the actions performed by individuals from video recordings.
In industrial environments, this capability is essential for ensuring adherence to standard operating procedures, promoting safe machinery usage, and safeguarding worker safety.
Existing approaches typically interpret observed behaviour using either rule-based reasoning or probabilistic models.
In this work, we focus on the latter and address real-world scenarios where the partial observability caused by inherent occlusions poses a significant challenge.
Most competitors~\cite{peixoto2001realtime,urgo2019human,andrianakos2019approach,urgo2024monitoring} rely on the interpretation of human actions, which can be ambiguous, deviate from standard procedures, or vary significantly across operators.
In contrast, our approach centers on detecting assembly components and then infers the underlying action by identifying the predefined assembly instruction that is most likely to have resulted in the observed configuration.

\noindent\textbf{Digital twins} are virtual replicas of physical objects that mirror their real-world counterparts either in appearance alone or also in functionality~\cite{jones2020characterising,liu2021review}.
They are essential for enabling the analysis and simulation of complex systems or processes, particularly when direct inspection is impractical, costly, or hazardous.
As a result, digital twins are playing an increasingly important role in manufacturing~\cite{cimino2019review,soori2023digital}, robotics~\cite{feddoul2023exploring,malik2024digital}, and autonomous driving~\cite{hu2023simulation,wang2024digital}.
More recently, they have become valuable tools for training, testing, and monitoring computer vision systems.
In particular, they have been used in combination with rendering engines~\cite{blender,denninger2019blenderproc} to generate synthetic datasets for training object detectors~\cite{hodan2019photorealistic}, pose estimators~\cite{labbe2022megapose}, and segmentation models~\cite{dosovitskiy2017carla}.
A key advantage of using digital twins in this context is the automatic generation of accurate annotations, reducing the burden of manually labelling real-world data.
In this work, we create digital twins of the assembly components, replicating both their shape and appearance.
We extract precise 3D geometry from synthetic CAD models and approximate visual appearance using synthetic materials~\cite{blenderkit,ldraw} that simulate photometric properties (e.g., colour, texture) and physical characteristics (e.g., reflectivity, signs of wear).

\noindent\textbf{Object detection} involves identifying and localizing objects in an image by predicting their categories and corresponding bounding boxes.
Key challenges include handling occlusions, scale variations, and ensuring temporal consistency when processing videos.
The field has evolved from traditional approaches based on hand-crafted features~\cite{dalal2005hog,felzenszwalb2009dpm} to data-driven approaches based on deep learning, which are generally classified into \emph{single-stage} and \emph{two-stage} methods.
Single-stage detectors~\cite{liu2016ssd,redmon2016yolo} perform object localization and classification jointly, offering faster inference.
In contrast, two-stage detectors~\cite{girshick2014rcnn,ren2015fasterrcnn} first generate region proposals identifying areas where objects may be located and then classify them, usually achieving higher accuracy at the cost of increased latency.
Models based on CNNs~\cite{liu2016ssd,redmon2016yolo} have faster training convergence and lower inference times, while methods based on Transformers~\cite{carion2020detr} enable end-to-end training and achieve better accuracy, but require higher computational costs.
In this work, we adopt YOLOv8~\cite{jocher2023ultralytics,yolov8}, a CNN-based single-stage detector that offers a good trade-off between accuracy and efficiency.
We finetune it on a synthetic dataset generated using task-specific digital twins and the BlenderProc~\cite{denninger2019blenderproc} rendering engine.

%% file: sections/3_method.tex
\section{Method}\label{sec:method}

\noindent\textbf{Assembly tasks} are formally defined by three components: \emph{objects}, \emph{configurations}, and \emph{actions}.
Objects include physical \emph{elements}, which are manipulated during the assembly process, as well as \emph{trays}, which are designated regions of the workspace associated with specific functions, such as storing elements before use, collecting unmounted elements, or holding the required tools.
Note that we do not require trays to be physical containers.
Configurations describe the relationships among objects and are defined by a set of Boolean predicates over one or more objects that hold true in a specific assembly state.
Actions represent admissible interactions with objects that cause transitions between configurations by modifying the associated predicates.
Every action is associated with a set of \emph{preconditions}, i.e. predicates that must be satisfied for the action to be applicable.
We model an assembly task as a state-transition system, where states correspond to configurations and transitions to actions~\cite{dean1991planning,ghallab2004automated}.
In this framework, a planner can identify the correct executions of an assembly task by finding ordered sequences of states transitioning from the input \emph{initial configuration} $C_\text{init}$ to the target \emph{final configuration} $C_\text{final}$ through valid actions.
This system can be represented as a directed graph, where nodes correspond to configurations (with $C_\text{init}$ as root and $C_\text{final}$ as the sole leaf), edges represent actions, and paths connecting $C_\text{init}$ to $C_\text{final}$ define valid plans.
Note that multiple such paths may exist, each corresponding to a different valid plan (see Fig.~\ref{fig:qual_lego}, bottom).

\noindent\textbf{Visual monitoring} of an assembly task consists in estimating the current state of the system from visual data by observing its evolution during the execution of a plan.
The task of visual monitoring is difficult for two main reasons: observations provide uncertain information, and the system is only partially observable.
For this reason, we adopt a probabilistic approach~\cite{viterbi1967error} to find the sequence of states that maximizes the observations' probability.
The observations on the working environment are provided by the perception module in the form of object detections.
By processing these images, the system identifies and tracks the components of the assembly, enabling it to predict the probability distribution.
Given the current view of the scene, the system updates the probability of being in a specific configuration based on the observed state of the components.
By reasoning about the most probable configuration of the assembly, the system monitors the operator.
This enables accurate tracking of the assembly progress and facilitates reliable execution of the process, ultimately guiding it to the desired target configuration $C_\text{final}$.

\input{figures/diagram/diagram}

\noindent\textbf{System overview.}
Fig.~\ref{fig:diagram} shows the architecture of \acronym, comprising three main modules: \emph{digital twin} (green), \emph{perception} (pink), and \emph{reasoning} (azure).
The perception module processes multi-view video streams of an assembly task, crops predefined regions corresponding to various trays, and performs frame-level detection of assembly components within each region.
To ensure accurate detection of task-specific components, we finetune the object detector on synthetic data generated by the digital twin module.
Spatial and temporal coherence is enforced through multi-view fusion and temporal smoothing, respectively.
The reasoning module then uses the resulting detections to infer the most likely action being performed, based on a predefined set of assembly instructions.

\subsection{Digital twin}\label{sec:method_digitaltwin}

\noindent\textbf{Objects.}
Each element involved in the assembly process is modeled using its corresponding synthetic 3D representation from the input CAD models.
Although geometric information (i.e., their 3D shape) is directly derived from these models, the visual appearance is obtained by assigning the appropriate textures and materials~\cite{ldraw,blenderkit}, designed to resemble the physical counterparts of the elements.

\noindent\textbf{Scene generation.}
To simulate realistic assembly environments, we procedurally generate 3D scenes by placing the elements on a planar surface.
The placement is randomized but guided by prior knowledge and by preventing spatial conflicts among objects.
This allows the generation of physically plausible and diverse scenes that reflect the variability observed in real assembly processes.

\noindent\textbf{Synthetic dataset generation.}
We use a photorealistic rendering engine to generate images capturing partial views of the 3D scene~\cite{blender,denninger2019blenderproc}.
To increase data variability, we perform random selection of camera viewpoints, distances from the scene, and lighting conditions.
We generate photometric information by capturing RGB channels and depth information by capturing range distances from the sensor.
In addition, we render the foreground object as a distinct layer to obtain segmentation masks, which are used to extract bounding boxes to supervise the training of our object detection model.

\subsection{Perception}\label{sec:method_perception}
The perception module takes as input RGBD frames captured from multiple calibrated viewpoints and produces as output the assembly elements detection. 
\noindent\textbf{Tray localization.}
The trays are positioned at pre-defined locations that do not change during the assembly process.
While their positions are fixed, the elements within each tray are placed freely, without any specific ordering or categorization.

\noindent\textbf{Object detection.}
We train an object detector $\Phi_\Theta$ on a synthetic dataset generated using a digital twin of the task-specific assembly environment (see Section~\ref{sec:method_digitaltwin}). 
At inference time, $\Phi_\Theta$ performs object detection frame-wise, producing 2D bounding boxes, class predictions, and confidence scores.

\noindent\textbf{Multi-view fusion.}
To improve the robustness and accuracy of  $\Phi_\Theta$, we leverage multiple cameras  through a multi-view fusion strategy. The process begins by selecting, for each viewpoint, the 2D detections that fall within each predefined tray region. This results in a set of per-tray, per-view detections with associated confidence scores.
Every pixel inside the 2D detection is back-projected using depth, intrinsics, and the pinhole model, then transformed to the global frame via extrinsics, yielding a 3D point cloud per detection per view. To associate detections across viewpoints, we match these point clouds using a pairwise Point Cloud Intersection-over-Union (IoU) metric.
This is based on a radius-based intersection strategy. Formally, given two point clouds $\mathcal{P}_1$ and $\mathcal{P}_2$, and a radius threshold $r$, we define the intersection set as: $\text{Intersection}(\mathcal{P}_1, \mathcal{P}_2, r) = \left\{ p \in \mathcal{P}_2 \;\middle|\; \exists q \in \mathcal{P}_1 \text{ such that } \|p - q\| < r \right\}$. The corresponding IoU value is computed using the usual formula.
This method allows us to robustly identify overlapping detections across views, even in the presence of minor inaccuracies.
Once the pairwise IoUs are computed and matching detections are identified, the results are aggregated across all views to obtain a consolidated observation, i.e., a vector representing the presence and confidence of each element in the workspace. The multi-view fusion helps mitigate occlusions and provides a more reliable perception of the scene.

\noindent\textbf{Temporal smoothing.}  
To ensure consistency over time and reduce noise in the predictions, we apply a temporal smoothing strategy to the consolidated observation vectors. At each time step, the current observation vector is combined with the previously smoothed one using a weighted average. The weighting is asymmetric: if the current score of an element is higher than the previous one, it is given more weight to allow rapid adaptation to improvements; otherwise, the previous score is favored to dampen abrupt drops.

\subsection{Reasoning}\label{sec:method_reasoning}

\noindent\textbf{Assembly instruction formalization.}
Together with domain experts, we formalize the description of system configurations, the set of actions that can be performed and, for each action, the set of preconditions that have to hold in order to perform it.
A configuration involves several objects and a set of predicates that can hold among them.
In the \lego scenario, for example, we identified 13 objects ($ E_1, E_2, E_3, E_4, E_5, \allowbreak E_6, E_4', E_5', E_6', T_{\text{in}}, T_{\text{out}}, T_{\text{work}}, T_{\text{aux}}$) and 6 predicates (\scalebox{1.0}[1.0]{\tt{is\_joined}}, \scalebox{1.0}[1.0]{\tt{is\_split}}, \allowbreak \scalebox{1.0}[1.0]{\tt{is\_mounted}}, \scalebox{1.0}[1.0]{\tt{is\_accessible}}, \scalebox{1.0}[1.0]{\tt{is\_free}}, \scalebox{1.0}[1.0]{\tt{do\_contain}}).

A state (or configuration) is specified by the set of predicates that hold when the system is in that state.
Actions describe the different modes the systems can be modified to pass from one configuration to another. Their effect is to update the list of predicates that are valid before the execution of the action.
In the \lego scenario, the set of elementary actions is $A = \{ \scalebox{1.0}[1.0]{\tt{join}}, \scalebox{1.0}[1.0]{\tt{split}}, \scalebox{1.0}[1.0]{\tt{mount}}, \scalebox{1.0}[1.0]{\tt{remove}}, \scalebox{1.0}[1.0]{\tt{put}}, \scalebox{1.0}[1.0]{\tt{take}}\}$.
In practice, the description of the reconfiguration process in terms of elementary actions often turns out to be redundant because some sequences of actions are always performed together and can be considered atomic.
Therefore, we have described the process as a set of steps, each defined by a sequence of actions, and the preconditions that must hold in order to be executed.
In this way, we obtain a more meaningful description of the activity to be performed by the operator, hiding too detailed and often obvious instructions.
For example, we defined the step \textit{Mount element $E_4$ on $E_3$} which is composed by two actions (\scalebox{1.0}[1.0]{\tt{take}}($E_4$, Tin) and \scalebox{1.0}[1.0]{\tt{mount}}($E_4$, $E_3$)) and three preconditions (\scalebox{1.0}[1.0]{\tt{do\_contain}}(Tin,$E_4$), \scalebox{1.0}[1.0]{\tt{is\_accessible}}($E_3$), \scalebox{1.0}[1.0]{\tt{is\_free}}($E_3$)).
In the \lego scenario we have identified 10 steps.

\noindent\textbf{State-graph generation.}
The state-transition system can be used to find a plan (i.e. a sequence of steps) that, from some given initial configuration, permits to achieve an objective, typically specified as a goal state.
By exploring the state-space starting from $C_\text{init}$, we automatically construct a directed graph $\mathcal{G}$ representing all feasible sequences of steps reaching $C_\text{final}$ (Fig.~\ref{fig:qual_lego}, bottom).

\noindent\textbf{Graph-based probabilistic reasoning.}
The control of the execution of a given plan is performed by a controller module by observing the system evolution during the plan execution.
In our case, the controller module does not take decisions, but operates as an external observer trying to estimate the current state of the system.
Obviously, if it detects deviations from a valid plan, a warning message can be sent to the operator.
Observations aim to test which predicates are valid in the observed configuration to provide the probability that the system is in that state.
Because observations are intrinsically partial and uncertain, we adopt a probabilistic approach trying to estimate the most probable sequence of steps performed (or traversed configurations) given all the observations from the beginning to the current time.
The Viterbi algorithm works by iteratively calculating the probabilities of being in each state at each time step, considering the previous state and the current observation: 
\[
V_{1,k} = P(y_1|s_k) \pi_k \hspace{2cm}
V_{t,k} = \max_{x \in S} (P(y_t | s_k) a_{x,s_k} V_{t-1,x})
\]
where $S=\{s_1,\ldots,s_n\}$ are the states, initial probabilities $\{\pi_1, \ldots \pi_n\}$ are all zero except for the initial state (set to 1.0), $y_t$ is the observation vector at time $t$, the transition matrix $a$ is estimated from $\mathcal{G}$ and $P(y|s)$, the probability of observing $y$ assuming the system is in state $s$, is estimated by comparing $y$ with its expected value $E_y(s)$, i.e. $P(y|s) \propto \mathrm{exp} \left( - \lVert y - E_y(s) \rVert / \sigma \right)$.
Estimates $V_{t,k}$ can be used to backtrack and find the most likely sequence of states (the Viterbi path).

%% file: figures/diagram/diagram.tex
\begin{figure}[t]
    \centering
    \begin{overpic}[trim=40 20 10 20, width=\textwidth]{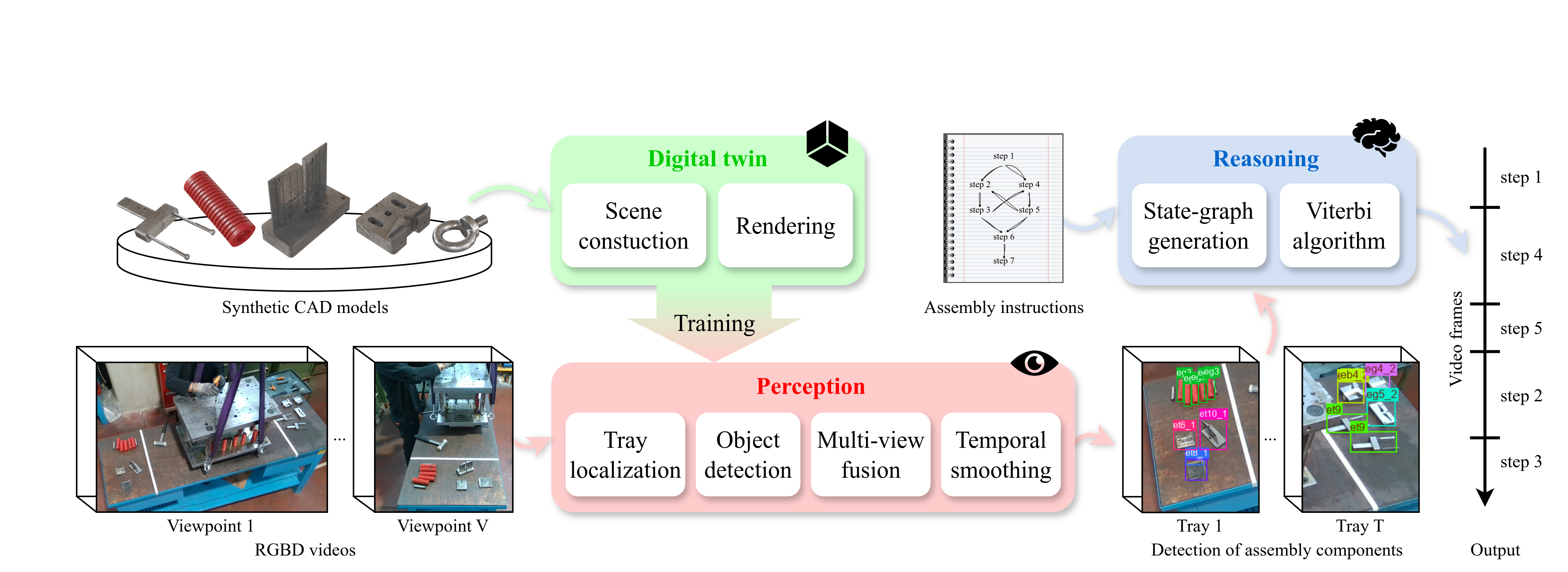}
    \end{overpic}

    \caption{
    Overview of \acronym.
    Multi-view video frames are processed by the perception module (pink) to detect assembly components using a detector trained on a synthetic dataset generated by the digital twin module (green).
    These detections, along with prior task knowledge (assembly instructions), are passed to the probabilistic reasoning module (azure) to estimate the action being performed.
    }
    \label{fig:diagram}
\end{figure}

%% file: sections/4_results.tex
\section{Results}\label{sec:results}

\subsection{Experimental setting}

We conduct experiments in two distinct scenarios: a controlled \lego scenario and an industrial scenario. In both scenarios, we use three RGB-D sensors for perception, and the communications between system modules are handled via sockets.
We train as object detector $\Phi_\Theta$, the pretrained YOLOv8-X model (68.2M parameters) and finetune it for 100 epochs.
We employ a cosine decay schedule, and the SGD optimizer.
Input images are resized to 480$\times$480 pixels with standard data augmentation techniques (random scaling, rotation, and HSV variations).

\noindent\textbf{Datasets.}
For the \lego scenario, we create a digital twin comprising the 13 unique objects.
The 3D models are constructed using Stud.io~\cite{studio} and Ldraw~\cite{ldraw} to ensure accurate representation.
We generate for this task a synthetic dataset of 10,000 images by using either Blender~\cite{blender} or BlenderProc~\cite{denninger2019blenderproc} as rendering engine.
The industrial scenario uses a more complex digital twin with 20 distinct objects.
These are modeled by extruding 2D blueprints to replicate the 3D geometry and textured using realistic materials from BlenderKit~\cite{blenderkit} to simulate surface appearance and signs of wear.
A similar BlenderProc pipeline is employed to generate approximately 25,800 synthetic images.
In both scenarios, real-world background images of the workspace are collected without any objects to detect.
These images are included during the training of the object detector.
This strategy helps to reduce false positive detections during inference.

\noindent\textbf{Metrics.} We evaluate object detection performance using mean average precision (mAP) on manually annotated real-world validation data.

\subsection{Results}

\noindent\textbf{Quantitative Results.}
We present quantitative evaluations of \acronym's object detection performance in both \lego and industrial scenarios. 
Tables \ref{tab:lego_obj_det} and \ref{tab:mds_obj_det} detail the average mAP results for each setting, highlighting the impact of data augmentation (Aug.) and background images (BG) during training. 
\input{tables/result_table_side}
\input{figures/lego_figures/lego_figure2}

\emph{\lego scenario.}
Tab.~\ref{tab:lego_obj_det} shows the average mAP across three test sequences. Our baseline model, trained solely on synthetic data generated with Blender (B) without augmentation or background, achieves 60.4\% mAP. Applying data augmentation improves the average to 71.8\%. Using BlenderProc (BProc) as the rendering engine further increases performance to 76.4\% mAP. Incorporating background augmentation yields the best results, reaching 79.7\% mAP overall.
Models trained exclusively on real images outperform those trained solely on synthetic data. However, these real-image-only models exhibit reduced generalization on new sequences with different camera setups.

\emph{Industrial scenario.}
Tab.~\ref{tab:mds_obj_det} shows the mAP results across the three cameras and their average.
In this more challenging scenario, our method achieves a peak performance of 58.4\%, reflecting the increased difficulty in detecting similar objects.
The models trained using only real images perform worse than in the \lego scenario, likely due to the limited data (only 8 images).
Tab.~\ref{tab:mds_obj_det} includes ablation studies, such as variations using synthetic and real data (S+R), and SAM-6D~\cite{lin2024sam} performance. 
Combining synthetic and real data for training (S+R) yields improved results, highlighting the benefit of including real data when available.
SAM-6D, a model-based zero-shot method, achieves lower performance, likely due to its poor calibration.

\noindent\textbf{Qualitative Results.}
Figures~\ref{fig:qual_lego} and~\ref{fig:qual_industrial} show the state probabilities over time for different videos.
Each graph compares the ground-truth states (pink) with \acronym's predictions (blue or green) and with predictions using ground-truth object detections as observation vectors (azure).
Precision and recall are computed by matching predicted and ground-truth active states at each time step.

\emph{\lego scenario.}
At the bottom of Fig.~\ref{fig:qual_lego}, the complete assembly path is shown, illustrating the system’s ability to track the progression from the initial (left) to the final configuration (right).
\acronym closely follows the ground-truth sequence, demonstrating strong temporal alignment.
Notably, predictions from the trained object detector closely match those derived from oracle detections, indicating that the model, trained solely on synthetic data, generalizes well to the real-world and provides reliable input for downstream reasoning.

\emph{Industrial scenario.}
In Fig.~\ref{fig:qual_industrial}, the state probabilities over time also follow the ground-truth trend, but with greater variability. This is expected due to the increased complexity of the environment and actions.
Nevertheless, the system still captures the overall sequence of states effectively.
In both scenarios, predictions often slightly anticipate the ground-truth annotations. This is likely due to the nature of many steps, where the configuration change (e.g., removing an object from a tray) occurs at the beginning of the step. Moreover, the industrial scenario yields lower precision and recall, likely reflecting the increased complexity and more challenging conditions compared to the controlled setting.

\input{figures/industrial_figures/industrial_diagram}

\noindent\textbf{Failure cases and limitations.}
A key limitation is the perception module, where missed or incorrect detections, especially for similar objects, impact downstream reasoning. 
The system also struggles to distinguish between actions with subtle differences. 

%% file: tables/result_table_side.tex
\begin{table*}[t]
\centering
\begin{minipage}[t]{0.44\textwidth}
\centering
\tabcolsep 3pt
\caption{Object detection performance in the LEGO setting.
}
\label{tab:lego_obj_det}
\begin{tabular}{rlcc|c}
    \toprule
    & Data type & Aug. & BG & mAP \\
    \toprule
    {\color{gray} \scriptsize 1} & B, Synth. &        &        & 60.4 \\
    {\color{gray} \scriptsize 2} & B, Synth. & \cmark &        & 71.8 \\
    {\color{gray} \scriptsize 3} & BProc, Synth.       & \cmark &        & 76.4 \\
    {\color{gray} \scriptsize 4} & BProc, Synth.    & \cmark & \cmark    & \textbf{79.7} \\
    \midrule
    {\color{gray} \scriptsize 5} & Real\footnotemark[1] & &        & 73.5 \\
    {\color{gray} \scriptsize 6} & Real\footnotemark[1] & \cmark & \cmark & 77.7 \\
    \bottomrule
\end{tabular}

\end{minipage}
\hfill
\begin{minipage}[t]{0.54\textwidth}
\centering
\caption{Object detection performance in the industrial scenario.} \label{tab:mds_obj_det}
\begin{tabular}{rlcc|ccc|c}
    \toprule
    \multicolumn{2}{c}{Variant} & Aug. & BG  & Cam0 & Cam1  & Cam2 &  Avg.\\
    \toprule
    {\color{gray} \scriptsize 1} & Synth.  & \cmark & \cmark & 44.2 & 65.4 & 56.8 & 55.8 \\
    {\color{gray} \scriptsize 2} & Synth.  & \cmark & \cmark & 41.4 & 72.9 & 59.5 & 58.4 \\
     \midrule
    {\color{gray} \scriptsize 3} & Real &        &        & 14.9 & 21.1 & 11.9 & 15.0\\
    {\color{gray} \scriptsize 4} & Real & \cmark &        & 26.9 & 43.0  & 26.7 & 30.9\\
    {\color{gray} \scriptsize 5} & Real & \cmark & \cmark & 17.3 & 29.8 & 29.1 & 26.3\\
    \midrule
    {\color{gray} \scriptsize 6} & S+R    & \cmark & \cmark & 56.5 & 60.5 & 61.5 & 60.0 \\
    {\color{gray} \scriptsize 7} & S+R\footnotemark[2]    & \cmark & \cmark & 60.7 & 79.4 & 63.3 & \textbf{66.7} \\
    \midrule
    {\color{gray} \scriptsize 8} & SAM-6D & - & - & 20.2 & 19.9  & 10.2  & 16.7 \\
    \bottomrule
\end{tabular}
\end{minipage}
\end{table*}
\footnotetext[1]{Results on a subset of test sequences due to training data overlap.}
\footnotetext[2]{Model pretrained on Open Image V7}

%% file: figures/lego_figures/lego_figure2.tex
\begin{figure}[t!]
    \centering
    \begin{subfigure}[b]{0.49\columnwidth}
        \centering
        \includegraphics[width=\linewidth]{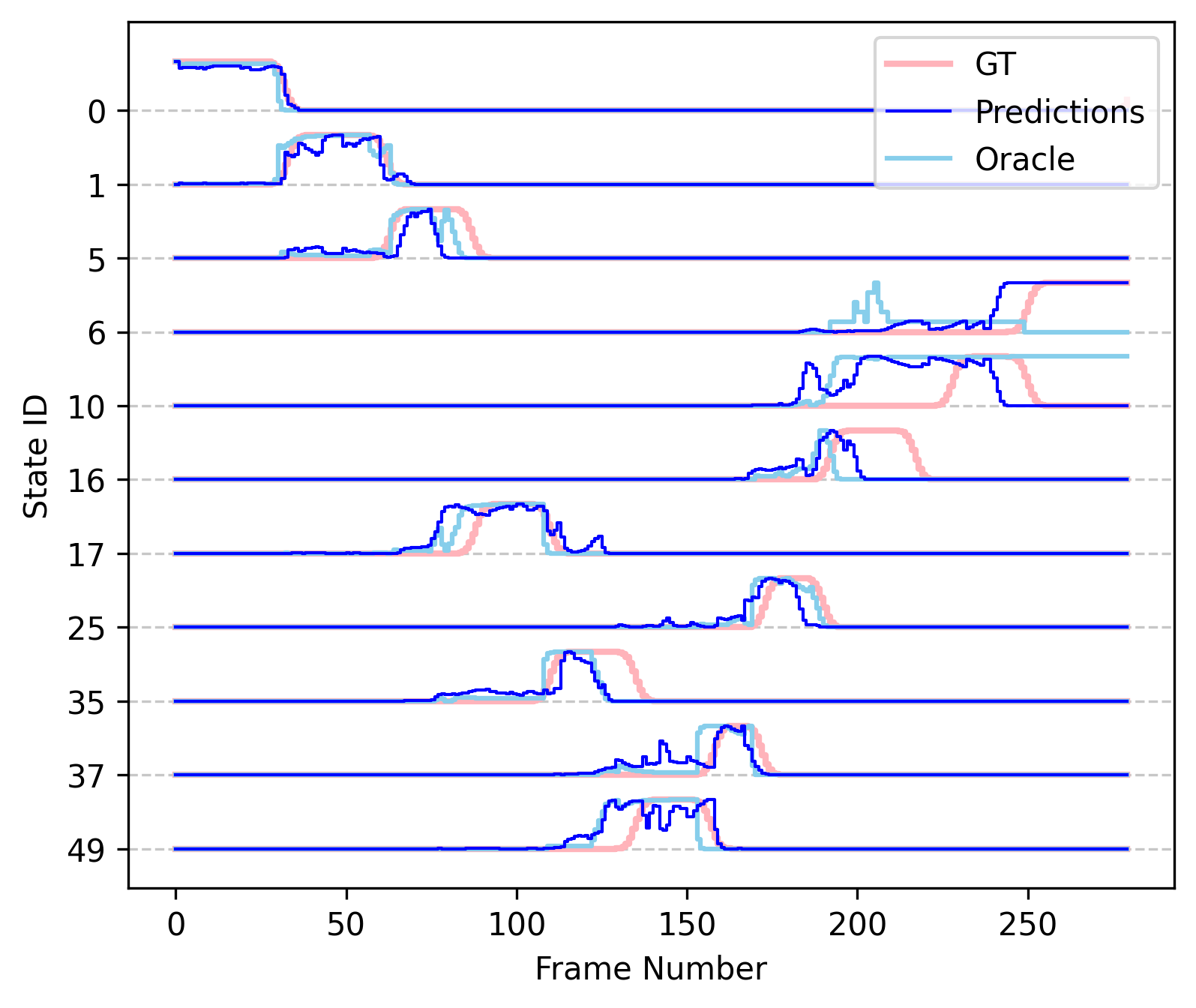}
        \label{fig:qual_lego_a}
    \end{subfigure}%
    \hfill
    \begin{subfigure}[b]{0.49\columnwidth}
        \centering
        \includegraphics[width=\linewidth]{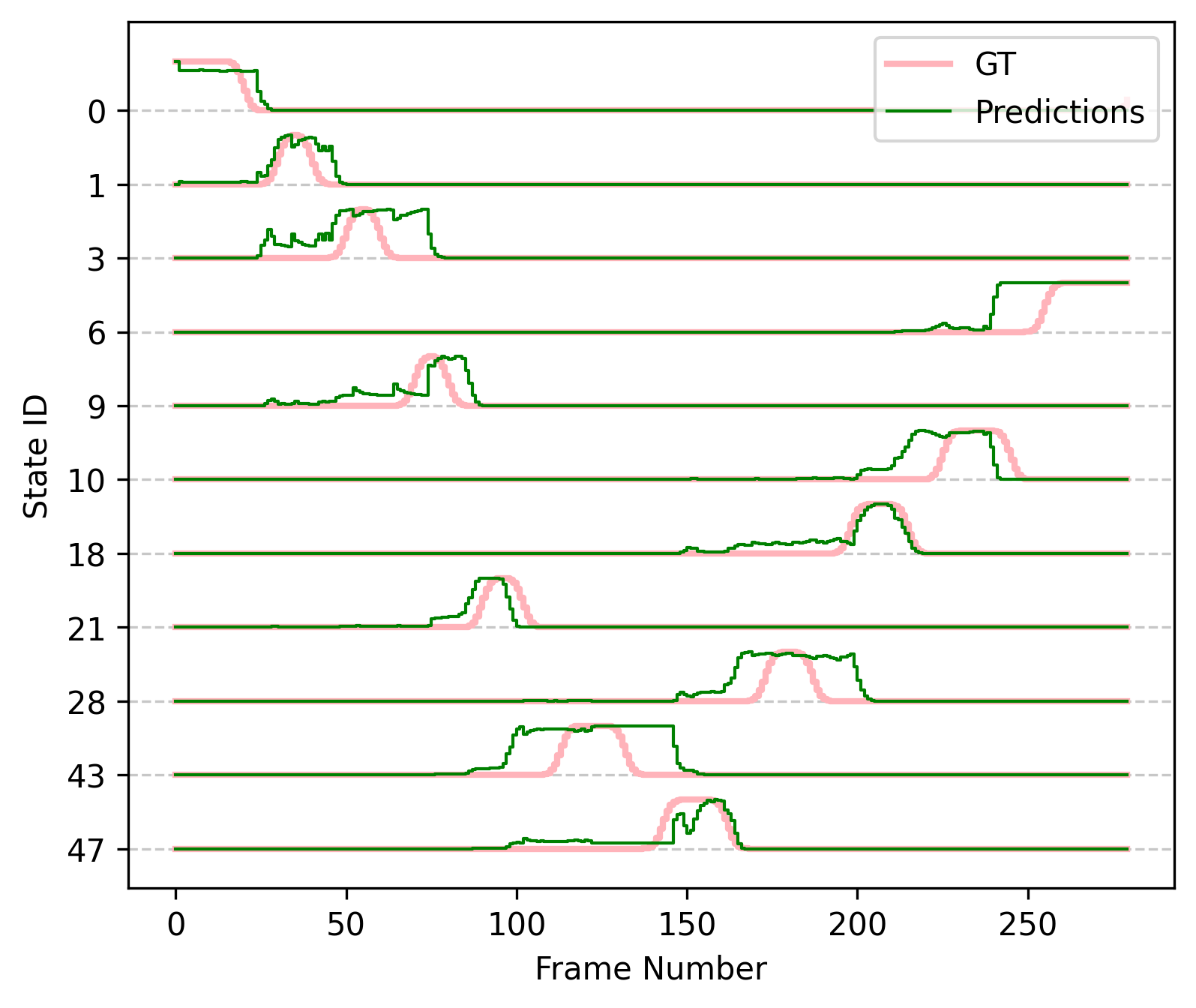}
        \label{fig:qual_lego_b}
    \end{subfigure}

    \vspace{-17.5mm} 

    \begin{minipage}[b]{\columnwidth}
        \centering
        \hspace{0.05\columnwidth}
        \includegraphics[trim=0 0 0 0, clip, width=0.45\linewidth]{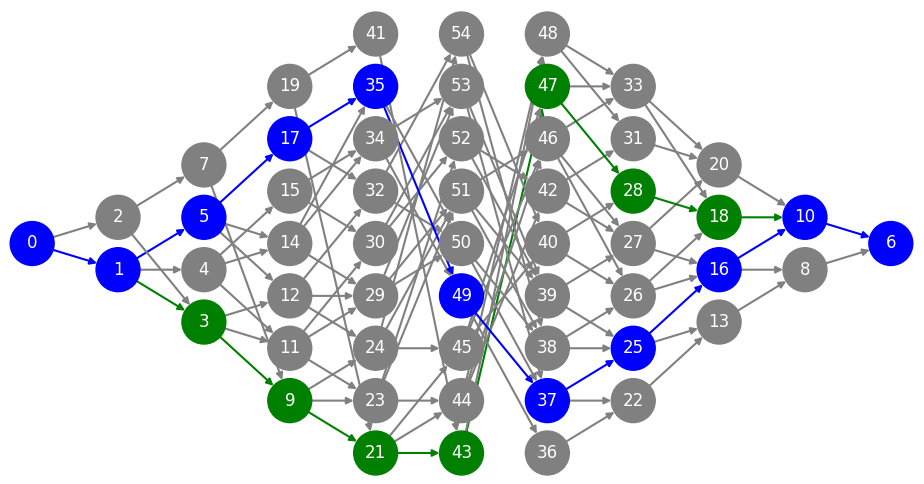}
        \label{fig:qual_lego_c}
    \end{minipage}

    \vspace{-3.5mm}
    \caption{Visual monitoring of the \lego assembly: Prec. = 72.9, Recall = 77.9.}
    \label{fig:qual_lego}
\end{figure}

%% file: figures/industrial_figures/industrial_diagram.tex
\begin{figure}[t!]
    \centering
    \begin{overpic}[trim=0 0 0 0, clip, width=\columnwidth]{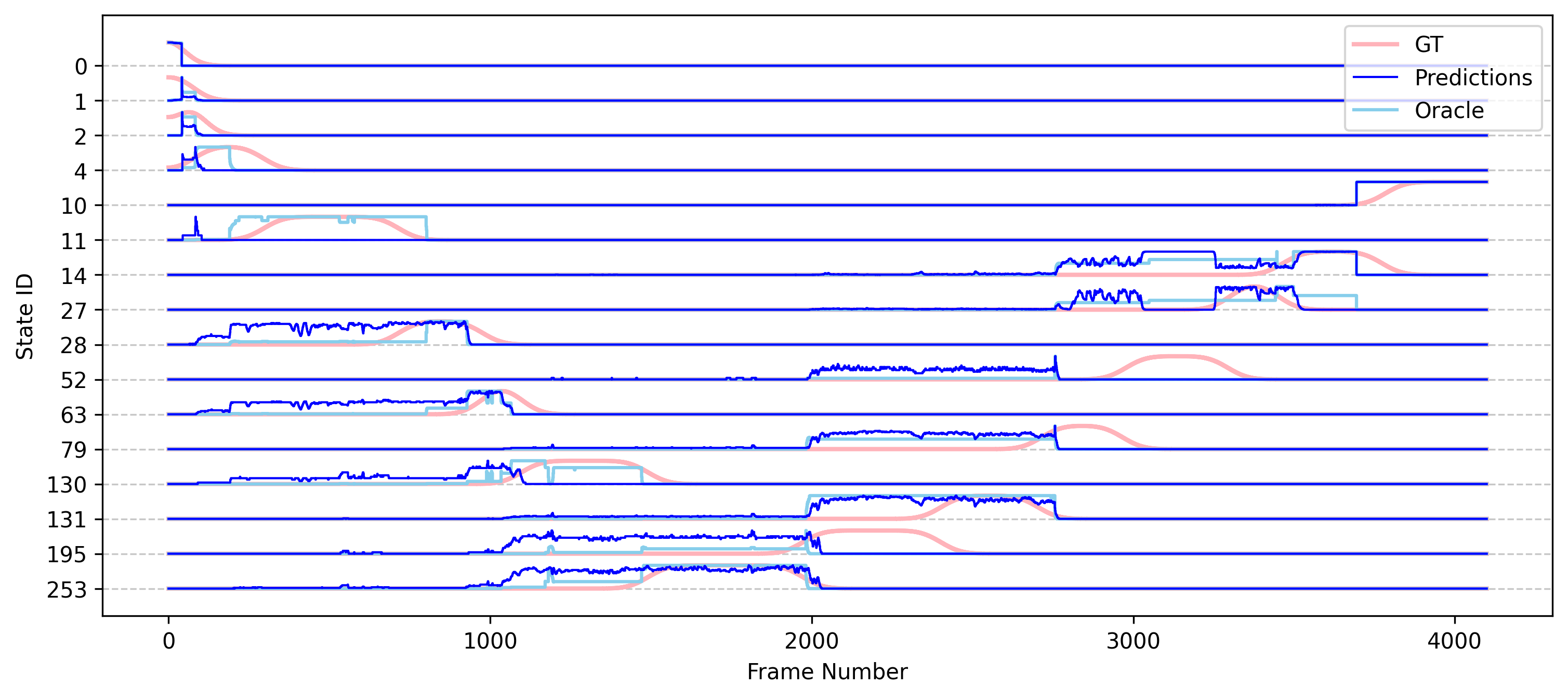}
    \end{overpic}

    \vspace{-4mm}
    \caption{
    Visual monitoring of the industrial assembly: Prec. = 43.1, Recall = 43.7.
    }
    \label{fig:qual_industrial}
\end{figure}

%% file: sections/5_conclusion.tex
\section{Conclusions}\label{sec:conclusions}

We presented \acronym, a real-time visual monitoring system for industrial assembly tasks that integrates AI-driven perception with probabilistic reasoning to infer assembly states from multi-view videos.
\acronym uses digital twins of the assembly components to finetune the object detector and predefined assembly instructions to guide reasoning.
As future work, we aim to incorporate hand pose estimation for finer action recognition and detect deviations from correct assembly procedures.

\noindent\textbf{Acknowledgments.}
This work has been partially funded by the Provincia Autonoma di Trento (Italy) under L.P. 6/99, as part of the NEXTMAG project.